\documentclass[11pt]{article}

\usepackage{graphics}
\usepackage[dvips]{graphicx}
\usepackage[left]{lineno}
\usepackage{multicol}
\usepackage{enumitem}
\usepackage{subfig}
\usepackage{color}
\begin{document}
\pagenumbering{gobble}

\vspace{1.2cm}

\begin{center}

{\bf  {\Large Understanding a Version of Multivariate Symmetric 
Uncertainty to assist in Feature Selection}}

\bigskip


{\small Gustavo Sosa-Cabrera$^a$\footnote{E-mail Corresponding Author: 
gdsosa@pol.una.py}, Miguel Garc\'ia-Torres $^b$, Santiago G\'omez$^a$, \\ 
Christian Schaerer $^a$ and Federico Divina $^b$
}
\smallskip

{\small
$^a$Polytechnic School, National University of Asunci\'on, P.O. Box 2111 
SL, \\San Lorenzo, Paraguay\\

$^b$Computer Science, Universidad Pablo de Olavide, ES-41013, Seville, 
Spain\\

}

\end{center}

\quad

\begin{abstract}

In this paper, we analyze the behavior of the multivariate symmetric 
uncertainty (MSU) measure through the use of 
statistical simulation techniques under various mixes of informative 
and non-informative randomly generated features.
Experiments show how the number of attributes, their cardinalities, 
and the sample size affect the MSU. We discovered a condition 
that preserves good quality in the MSU under different 
combinations of these three factors, providing a new useful 
criterion to help drive the process of 
dimension reduction.

\quad

{\footnotesize
{\bf Keywords}: feature selection, symmetrical uncertainty, 
multivariate prediction of response, high dimensionality.}
\end{abstract}

\quad

\textbf{1. Introduction}
\smallskip

There are fields, e.g. document processing and bioinformatics, 
in which multivariate datasets 
contain a huge amount of features and perhaps a low number 
of samples. 
In these spaces of high dimensionality, 
feature selection is a way  to  exclude those 
irrelevant and redundant features,  whose presence might 
complicate the task of knowledge discovery.

In classification tasks, a feature is considered 
irrelevant if it contains no information about the 
class and therefore it is not necessary at all for 
the predictive task. 
Besides, it is widely accepted 
that two features are redundant if their values are 
correlated.

There are several well known measures that compare features 
and determine their importance, such as the 
symmetrical uncertainty (SU)\cite{fayyad}. SU is a 
measure based on information that uses entropy and 
conditional entropy values to determine the correlation 
between pairs of features.
In order to consider  possible interactions that may exist among the 
different features,  the multivariate symmetric uncertainty (MSU)  
is proposed in \cite{Arias-Michel}, 
as a generalization of the bivariate measure. However, 
it is well known that in measures based on information, 
there is a bias in favor of those attributes with many 
values. In the following, we refer to the number of 
distinct labels of an attribute as its cardinality.

The aim of this paper is to analyze the MSU bias, considering 
the cardinalities of the attributes, the 
sample size and the size of the subset of attributes to 
be evaluated. 
To this aim, we used the Monte Carlo simulation technique 
to generate artificial data sets with informative and 
non-informative attributes with various numbers of values.

The rest of the paper is organized as follows.  
In \S 2 we provide the basic theoretical foundations. 
The experimental scenario is introduced in \S 3, 
whilst \S 4 provides the results of our experimentation. 
In \S 5 we draw the main conclusions and identify 
possible future developments.

\quad

\textbf{2. Theoretical foundations}
\smallskip

In this section we review some notions from Information 
Theory, that can be used in order to measure information as a 
reduction in uncertainty. 

The entropy $H$ of a discrete random 
variable $X$, with $\{x_1,\ldots,x_n\}$ as possible values and 
probability mass function $P(X)$, is a measure of the uncertainty 
in predicting the next value of $X$ and is defined as
{\small
\begin{equation}
H(X) := -\sum_{i} P(x_i)\log_{2}(P(x_i)),
\end{equation}}
\noindent where $H(X)$ can also be interpreted as a measure 
of a variety inherent to $X$, or the amount of information 
that is needed to predict or describe the outcome of $X$.

Given another discrete random variable $Y$, the conditional 
entropy $H(X|Y)$ quantifies the amount of information needed 
to describe the outcome of $X$ given that the value of $Y$ 
is known, and is defined as follows
{\small
\begin{equation}
	H(X|Y) := -\sum_{j} \left[ P(y_j)\sum_{i}P(xi|y_j) 
	\log_{2}(P(x_i|y_j)) \right], 
	\nonumber
\end{equation}} where $P(y_j)$ is the prior probability of 
the value $y_j$ of $Y$, and $P(x_i|y_j)$ is the posterior 
probability of a value $x_i$ for variable $X$ given that 
the value of variable $Y$ is $y_j$. 

Information Gain $(IG(X|Y))$ \cite{quinlan} of a variable 
$X$ with respect to a given variable $Y$ measures the 
reduction in uncertainty about the value of $X$ when the 
value of $Y$ is known, and is defined as 
{\small
\begin{equation}
IG(X|Y) := H(X) - H(X|Y). 
\end{equation}}

IG measures how much the knowledge of $Y$ makes the value of $X$ easier to predict, hence it can be used as a 
{\it measure of correlation}.
It can be shown that $IG(X|Y)$ is a symmetrical measure, 
which is a convenient property for a paired measure. 
However, IG presents a drawback: 
when $X$ and/or $Y$ have more values it is likely 
that they will appear to be {\it correlated}, hence IG 
tends to be larger when presented with attributes that have 
many different labels, that is, high cardinality. Definitions of cardinality will be given below. 

The IG values can be normalized 
using both entropies, originating the Symmetrical 
Uncertainty (SU) measure \cite{fayyad} expressed as 
{\small
\begin{equation}
	SU(X,Y) := 2 \left[ \frac{IG(X|Y)}{H(X) + H(Y)}\right].
\end{equation}}


The main limitation of $SU$ consists in taking into 
account only pairwise interactions and so it might 
lead to failure in the detection of redundancy when 
dealing with more than two features. 
To overcome this defect a Multivariate SU must be defined. To this 
end we use the total correlation definition
for $n$ variables \cite{wj,watanabe}
{\small
\begin{equation}\label{totalcorr}
	C(X_{1:n}) := \sum_{i=1}^{n} H(X_i) - H(X_{1:n}),
\end{equation}}
where
{\small 
\begin{equation}
	H(X_{1:n}) := H(X_1,...,X_n) := - \sum_{x_1} ... 
    \sum_{x_n} \\ P(x_1,...,x_n) 
    \log_2[P(x_1,...,x_n)]
\end{equation}}
is the joint entropy of the random variables 
$X_1, ..., X_n$.  

Based on the total correlation (\ref{totalcorr}), 
the Multivariate Symmetrical Uncertainty (MSU) is 
formulated as a generalization of the SU aimed to 
quantify the redundancy (or dependency) among more 
than two features \cite{Arias-Michel}. In this paper 
we use the following definition of MSU $\in [0,1]$ 
(for details see \cite{Arias-Michel}):
{\small
\begin{equation}
	MSU(X_{1:n}) := \frac{n}{n-1} \left[ \frac{C(X_{1:n})}
    {\sum_{i=1}^{n} H(X_i)}\right].
\end{equation}}


The cardinality measure can be used in order to define the amount of labels that can be releted to 
 a specific feature. 
The cardinality can be considered with respect to a 
single attribute (univariate) or with respect to 
several attributes including the class (multivariate). 
To formalize this concept we introduce the following 
definitions of cardinality. 

\vspace{0.1in}
{\it Definition 1}. Given a discrete or categorical 
attribute $A$, its {\it Univariate Cardinality}, 
denoted by $\left\vert{A}\right\vert$, is the number 
of possible distinct labels of $A$.

\vspace{0.1in}

{\it Definition 2}. Given a set of discrete or 
categorical attributes $A_1$, $A_2$, ..., $A_n$, $Y$, 
where $Y$ is a class feature, its 
{\it Multivariate Cardinality} is the number of possible 
label combinations among all features, including the class.

\vspace{0.1in}
Definition 1 tells us how diverse are the labels in a specific attribute. On the other hand, Definition 2 establishes how many combinations of labels are possible, measuring the diversity of information in the set. 

In the next section, we present the experiment setup 
and how the cardinality is used in the analysis. 

\quad

\textbf{3. Experimental scenario}
\smallskip

So as to generate the artificial datasets used in  the experimentation 
presented in this paper, we adopted the Monte Carlo simulation 
technique of White and Liu \cite{liu} including informative 
attributes made by Kononenko's 
method \cite{kononenko}.
The generated datasets present the following characteristics:
\begin{enumerate}
\setlength\itemsep{0em}
\item A classification attribute (``the class'') with either 
$2$ or $10$ possible values.
\item Informative and non-informative attributes 
with possible univariate cardinalities of 
$2$, $4$, $5$, $8$, $10$, $16$, $20$, $30$, $32$, $40$ or $64$.
\item Non-informative attributes were randomly generated from the 
uniform distribution independently of the class.
\item Informative attributes are made equally informative using 
Kononenko's method.
\item Attributes are made informative through their interaction 
by combining them through the $exclusive\ or$ function where 
noise is induced by 
$P(class = XOR(f_1,f_2)) = 0.95,P(class \not= XOR(f_1,f_2)) = 0.05$. 
\item  In one set of experiments, a fixed sample size 
of $1000$ and $5000$ cases was used and in the other ones the 
number of cases is allowed to vary.  
\end{enumerate}

The results of each experiment were averaged over $1000$ trials.

\quad

The Kononenko's method allows for equally informative multi-valued 
attributes. Thus, characteristic 4 is achieved by joining the values of 
the attributes into two subsets, the first one with 
$\left\lbrace1, ..., (V\;div\;2)\right\rbrace$ and the second one 
with $\left\lbrace (V\:div\;2\;+\;1), ..., V \right\rbrace$ for an 
attribute that has $V$ values. 
The probability that the value is from a given subset  depends on
the class, while the selection of one particular value inside a 
subset is random from the uniform distribution. 
The probability that 
the attribute's value is in one of the two subsets is defined by:
{\small
\begin{equation}
  P\bigg(j \in \left\lbrace1,...,(\lfloor\frac{V}
  {2}\rfloor)\right\rbrace|i\bigg) :=
      \left\{ \begin{array}{ll}
              1/(i+kC)    & \mbox{if $i$ mod 2 = 0} \\
              1 -1/(i+kC) & \mbox{if $i$ mod 2 $\ne$ 0}
               \end{array} \right.
\end{equation}}
where $C$ is the number of class values, $i$ is an integer indexing 
the possible class values $\left\lbrace c_1,...,c_i\right\rbrace$, 
and $k$ determines how informative the attribute is. A higher value 
of $k$ indicates a stronger level of association between the attribute 
and the class, so it makes the attribute more informative. All 
experiments in this work use $k = 1$.

With this scenario, we can now pursue our objective of analyzing the
MSU behavior based on the cardinalities of the attributes, the 
sample size and the size of the subset of attributes to be evaluated.

\quad

\textbf{4. Results}
\smallskip

In this section we will present the results achieved through 
various simulations on the artificially generated data sets. 

\begin{figure}[!htb]
    \subfloat[Informative and non-informative features without
    interaction. The cardinality of the class is $10$ and 
    sample size is $1000$ instances. 
    \label{subfig:a}]{
      \includegraphics[width=0.5\textwidth]{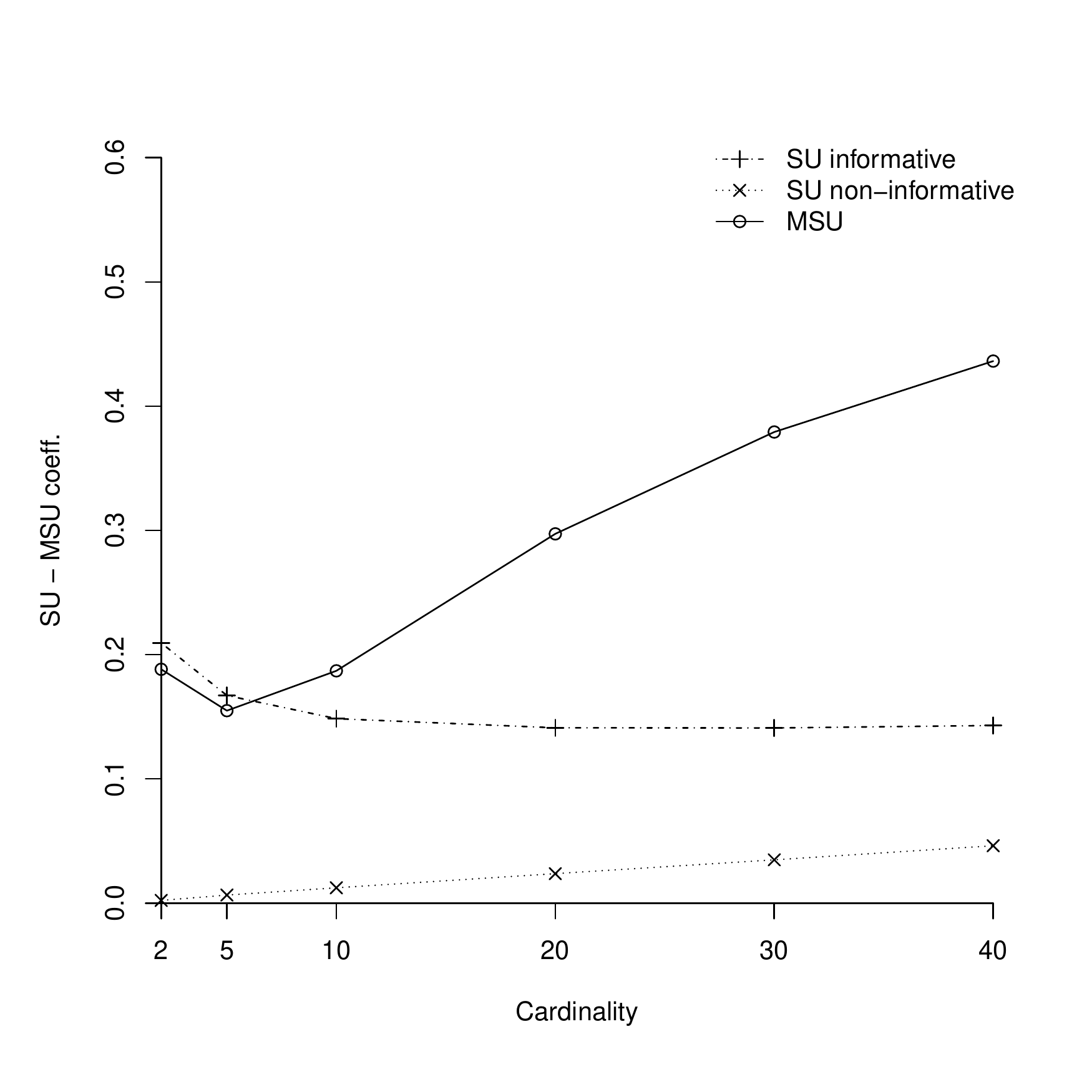}
    }
    \quad
    \subfloat[Interaction among XOR features with a noise 
    of $5\%$. The cardinality of features and the class 
    is $2$. \label{subfig:b}]{
      \includegraphics[width=0.5\textwidth]{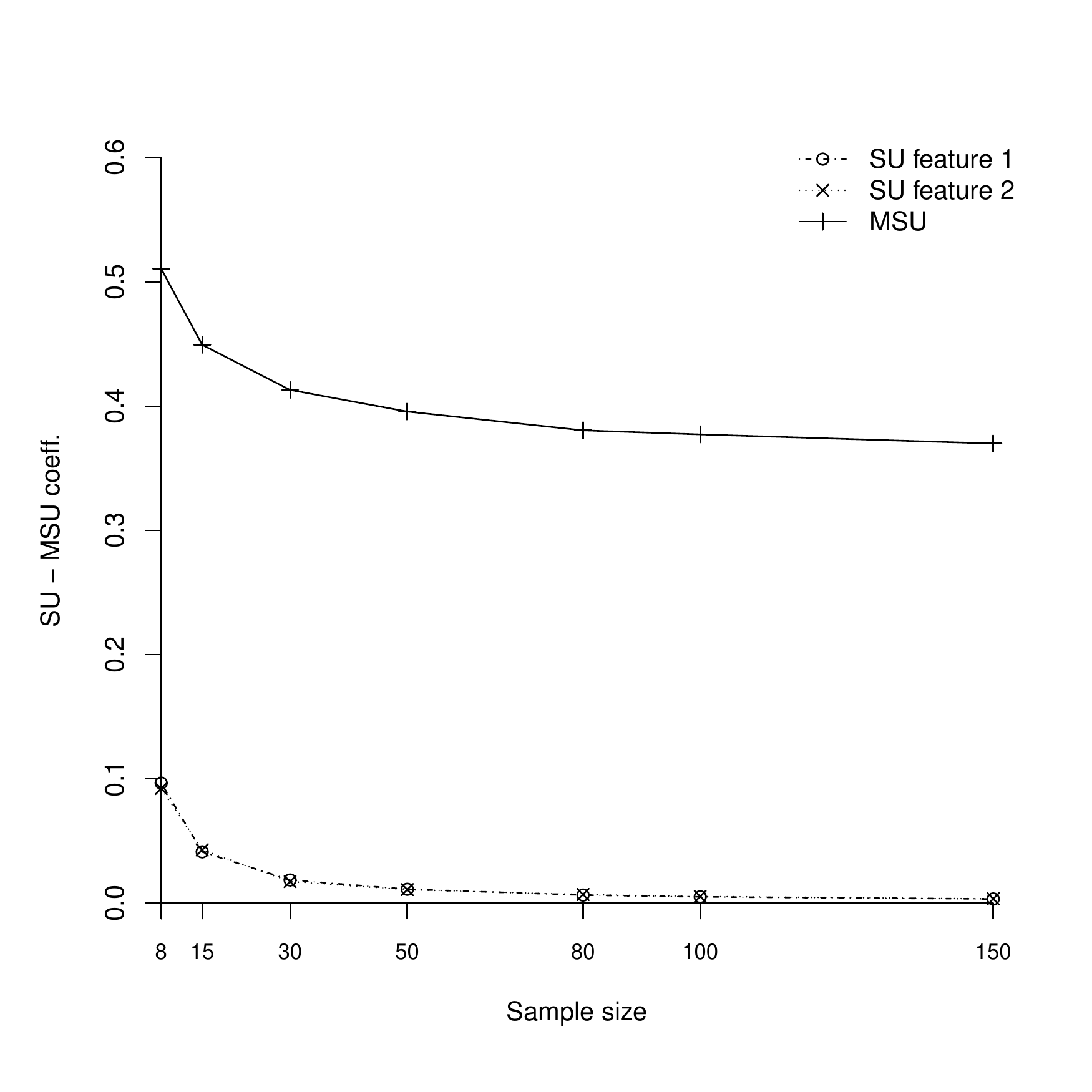}
    }
    \caption{The effects of varying the cardinalities and 
             sample size on SU and MSU.}
    \label{fig:figuras-ab}
\end{figure}

\textit{MSU with informative and non-informative variables.} 
Previous papers on the behavior of the SU have shown that 
an increase in univariate cardinality of the attribute 
produces a slow exponential-like decrease in SU if the 
attribute is informative, and a linear increase if the 
attribute is non-informative \cite{hall}, which we verify 
again. For the MSU, the interaction of these types of 
attributes renders an initial decrease followed by a steady 
growth as shown in Figure \ref{fig:figuras-ab}(a).
{\it MSU detects more information than SU.} 
In Figure \ref{fig:figuras-ab}(b), the SU graphs for 
features $f_1$ and $f_2$ (each having cardinality 2) 
overlap because they are equally uncorrelated to 
the class when on their own. Jointly taken though, 
they have a good correlation with the class which 
was made on purpose by $XOR$ing the features. 
The figure shows an important limitation of 
SU, since it  only considers one feature with 
the class: SU cannot detect situations where the 
conjunction of features is informative, 
as in this example. The MSU overcomes this limitation; 
as sample size grows the {\it variety} of the set of 
cases tends to stabilize and so does the MSU.
{\small 
\begin{table}[!htb]
  \centering
  \subfloat[][$MSU=0$]{
    \label{tab:a}
      \begin{tabular}{c|c||c}
          $f_1$ & $f_2$ & $class$\\
          \hline
          $b$ & $s$ & $p$ \\
          $b$ & $s$ & $q$ \\
          $b$ & $t$ & $p$ \\
          $b$ & $t$ & $q$ \\
          $a$ & $s$ & $p$ \\
          $a$ & $s$ & $q$ \\
          $a$ & $t$ & $p$ \\
          $a$ & $t$ & $q$ \\
      \end{tabular}
    }
    \qquad
  \subfloat[][$MSU=0.10$]{
    \label{tab:b}
      \begin{tabular}{c|c||c}
          $f_1'$ & $f_2$ & $class$\\
          \hline
          \boldmath$a$ & $s$ & $p$ \\
          $b$ & $s$ & $q$ \\
          $b$ & $t$ & $p$ \\
          $b$ & $t$ & $q$ \\
          $a$ & $s$ & $p$ \\
          $a$ & $s$ & $q$ \\
          $a$ & $t$ & $p$ \\
          $a$ & $t$ & $q$ \\
      \end{tabular}
    }
    \qquad
  \subfloat[][$MSU=0.18$]{
    \label{tab:c}
      \begin{tabular}{c|c||c}
          $f_1''$ & $f_2$ & $class$\\
          \hline
          \boldmath$c$ & $s$ & $p$ \\
          $b$ & $s$ & $q$ \\
          $b$ & $t$ & $p$ \\
          $b$ & $t$ & $q$ \\
          $a$ & $s$ & $p$ \\
          $a$ & $s$ & $q$ \\
          $a$ & $t$ & $p$ \\
          $a$ & $t$ & $q$ \\
      \end{tabular}
    }
    \caption{Cardinalities effects on MSU of non-informative 
    features.}
\end{table}}

\textit{Exploring how to set the sample size.} 
In this 
study we have identified a series of tendencies 
that make it necessary to establish the concept of 
cardinality more precisely. Thus we refer to univariate 
and multivariate cardinalities as specified in 
Definitions 1 and 2, respectively.

From here, the multivariate cardinality of a set of $n$ features 
including the $class$ is given by $\left\vert{class}\right\vert
\prod_{i=1}^{n} \left\vert{f_{i}}\right\vert$ 
where $\left\vert{class}\right\vert$ and 
$\left\vert{f_i}\right\vert$ are the univariate cardinality of the 
$class$ and of feature $f_i$, respectively. 

Figure \ref{fig:figuras-ab}(b) provides another very 
interesting point. The class in this example is a 
function of the two features $f_1$ and $f_2$, except 
for noise of up to $5\%$, and the good correlation 
pushed the MSU curve to relatively high values. 
When there is correlation among the variables, 
well-behaved MSU values tend to stabilize and the 
curve becomes nearly horizontal for all larger 
sample values. A negligible variation in the MSU means 
that a larger sample does not add significant 
information any more, that is, it is not necessary 
to continue increasing sample size.
In fact, we may 
establish a simple rule such as: ``stop increasing 
sample size whenever the variation in MSU is less 
than $0.01$". 
This actually occurs at sample size $80$, 
where the experimental MSU value went from $0.015$ 
down to $0.003$. 
\begin{figure}[!htb]
    \subfloat[Univariate and Multivariate Cardinality 
              mapping for informative features.
      \label{subfig:c}]{%
      \includegraphics[width=0.5\textwidth]{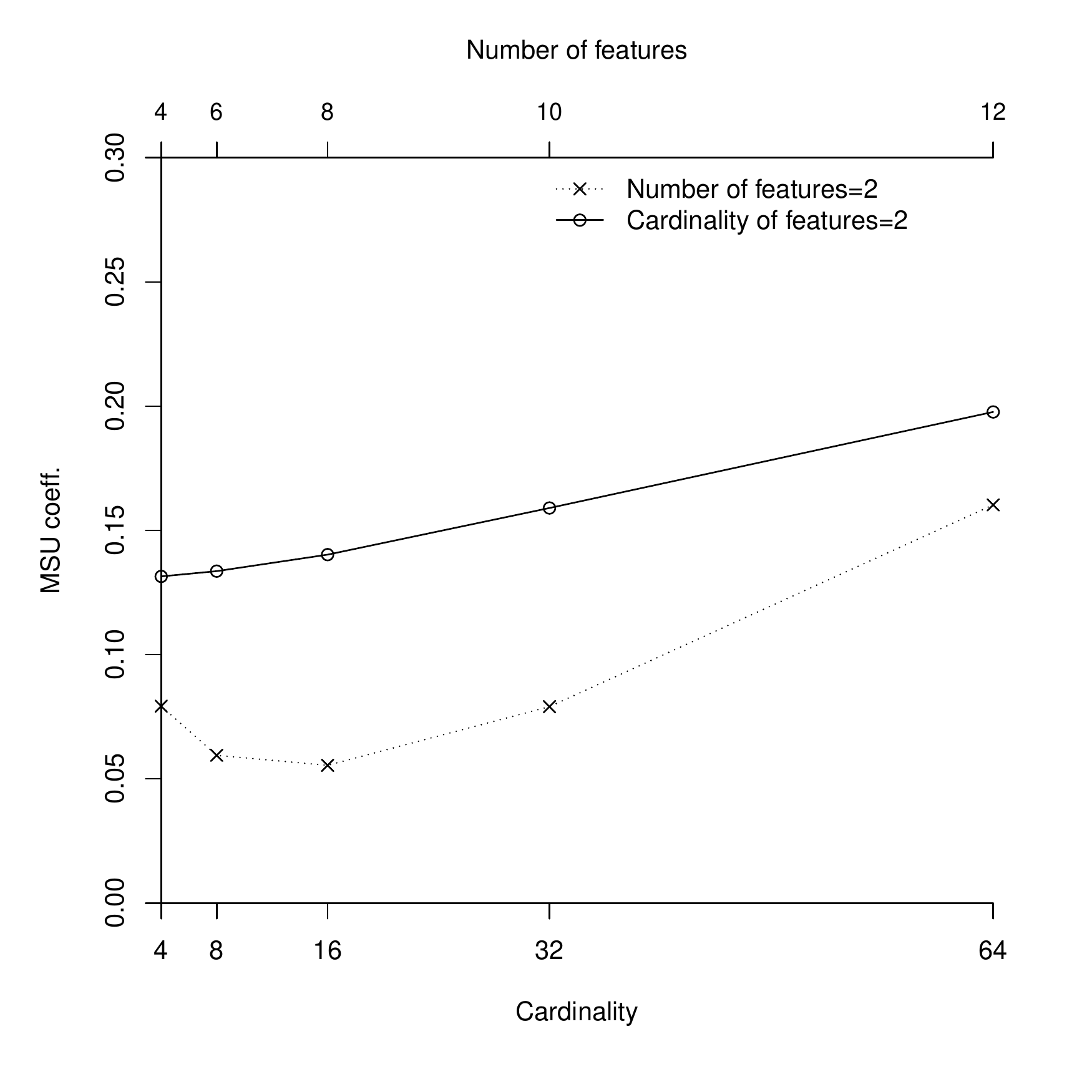}
    }
    \quad
    \subfloat[Univariate and Multivariate Cardinality 
              mapping for non-informative features. 
      \label{subfig:d}]{%
      \includegraphics[width=0.5\textwidth]{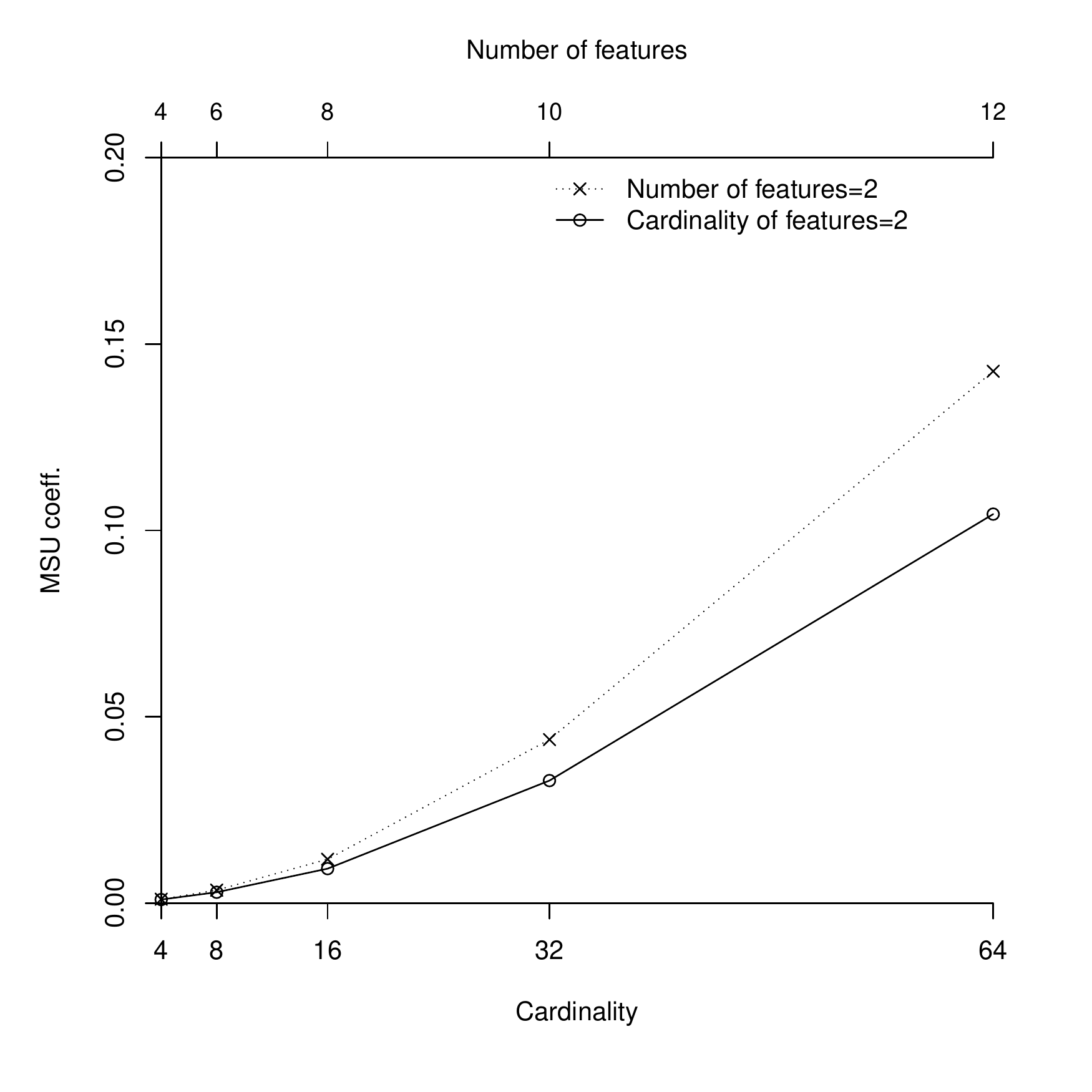}
    }
    \caption{The effects of varying the univariate and multivariate 
             cardinalities on MSU. Class cardinality is $2$ and  
             sample size is $5000$ instances.}
    \label{fig:figuras-cd}
\end{figure}

Since all variables have cardinality $2$, 
from the above their multivariate cardinality 
is $2*2*2 = 8$. 
A sample of just $8$ cases will hardly contain the $8$ 
different combinations; but a sample size of $80$ ($10$ 
times the multivariate cardinality of the set) is 
likely to capture enough information about any 
existing correlation.

\textit{Do cardinalities matter?}
Results clearly 
show that the tendency or bias of the MSU for 
non-informative features is conditioned by both 
the univariate and the multivariate cardinalities.

Table (\ref{tab:a}) displays the MSU of 
features $f_1$ and $f_2$ that were created randomly 
and independently from the $class$. The univariate 
cardinalities for both features and the class is $2$ .
In Table (\ref{tab:b}) one can appreciate 
how MSU increases when the value of a randomly picked 
instance is inverted (because multivariate cardinality 
is altered), and in Table (\ref{tab:c}) the effect 
of altering both cardinalities is shown.
\begin{figure}[htb]
    \subfloat[Using arbitrarily fixed sample sizes on 
              sets of informative and non-informative features. 
              The cardinality of the features is $2$. 
              \label{subfig:e}]{
      \includegraphics[width=0.5\textwidth]{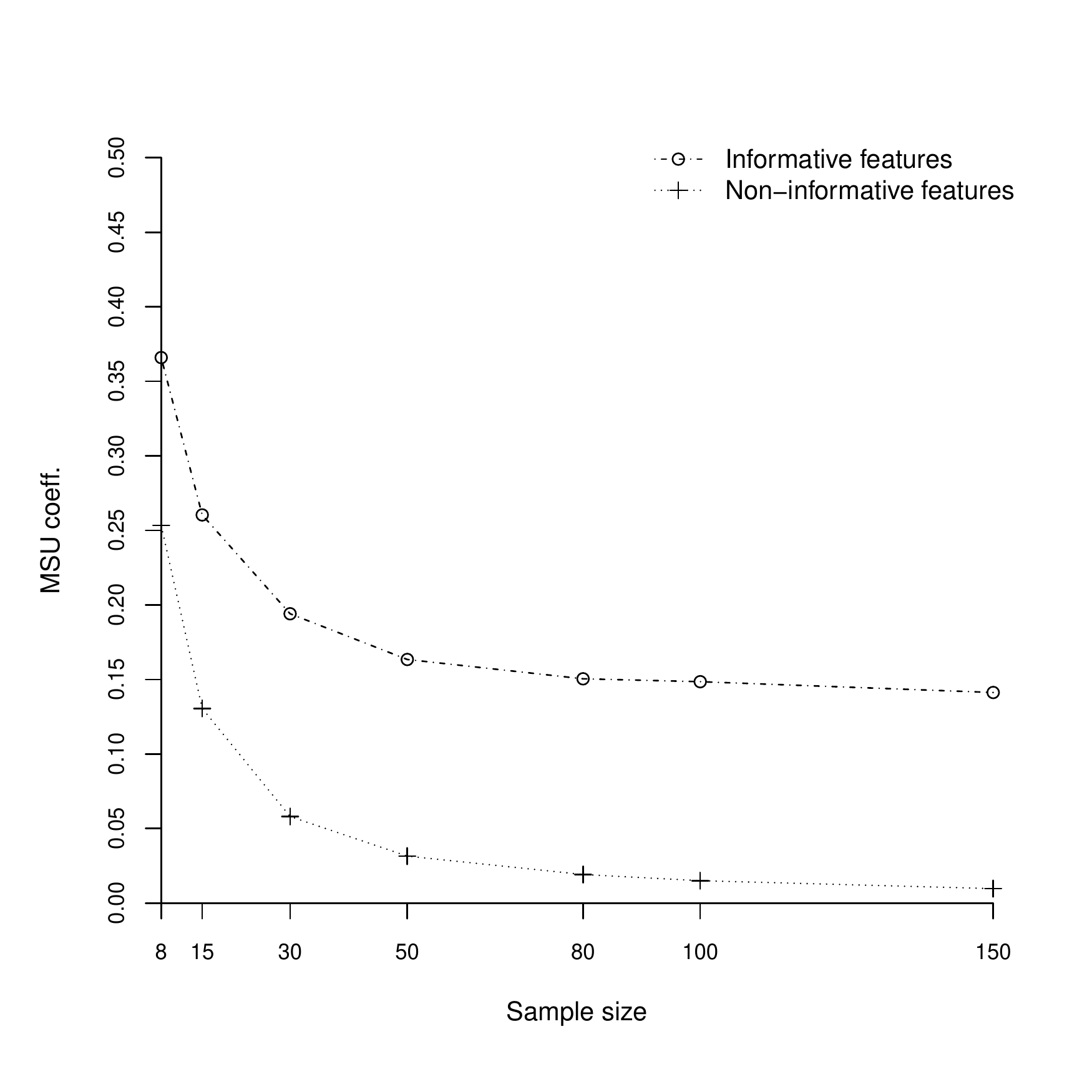}
    }
    \quad
    \subfloat[Effects of calculated sample size and the univariate 
              cardinality of features on sets of 
              informative and non-informative features.
              \label{subfig:f}]{%
      \includegraphics[width=0.5\textwidth]{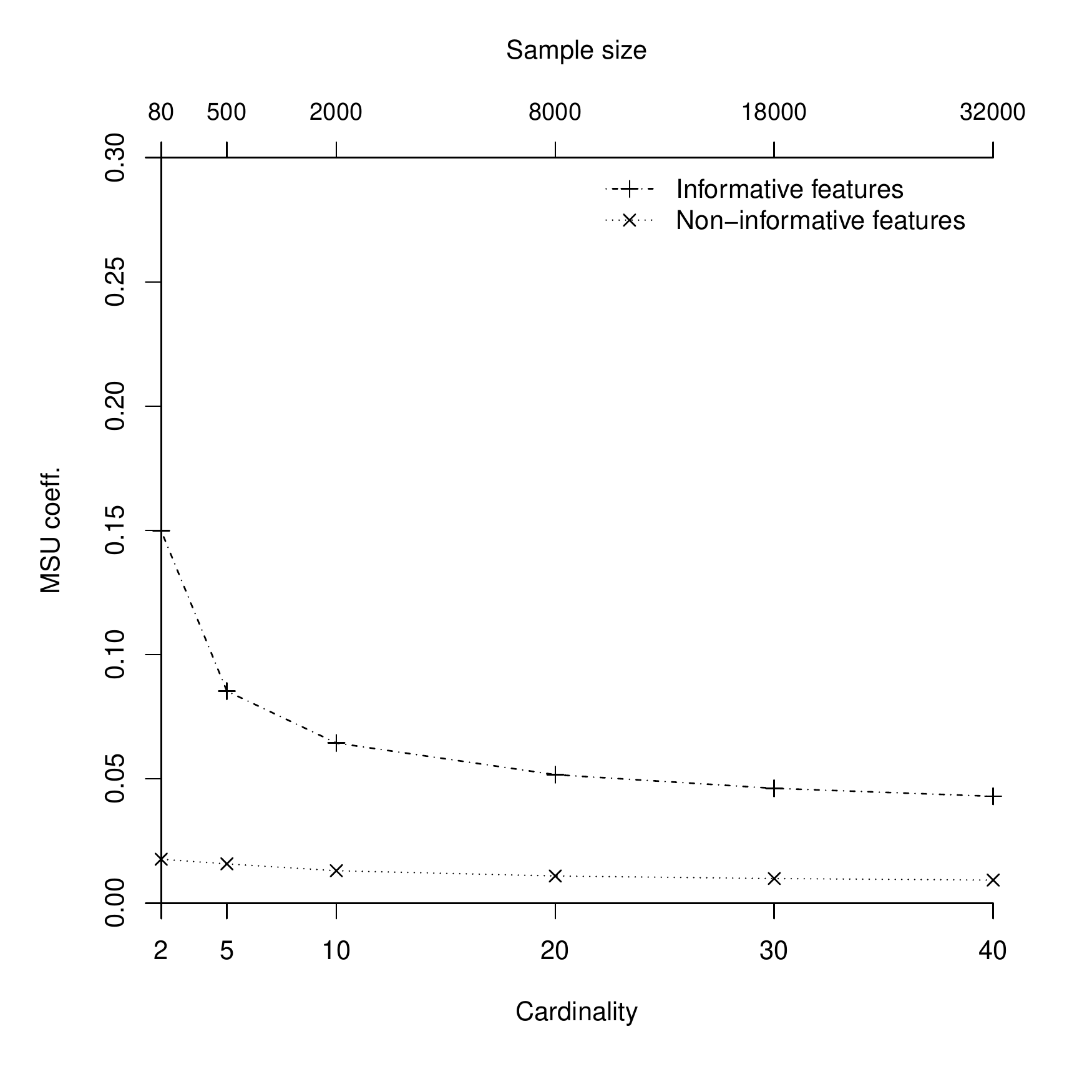}
    }
    \caption{The effects of varying sample size and 
             the univariate cardinality on MSU. 
             The cardinality of the class is $2$.}
    \label{fig:figuras-ef}
\end{figure}

The next question we address is the following: can a low number of features ($2$ for instance) 
with high cardinality yield MSU values comparable 
to a high number of features with low cardinality 
($2$ for instance)? 
Let us now keep the cardinality of 
the class at $2$ under a fixed sample size of $5000$. Results are shown in Figure \ref{fig:figuras-cd}(a) 
for informative features, and in Figure \ref{fig:figuras-cd}(b) 
for the non-informative ones. 
The dotted lines 
correspond to the MSU for $2$ features, with cardinalities 
from $4$ to $64$ (increasing univariate cardinalities); 
and the continuous lines represent the MSU from $4$ to $12$ 
features, each with cardinality $2$ (increasing multivariate 
cardinality). All curves show that higher univariate 
or multivariate cardinalities will produce higher MSU 
values, whether the set of features is informative or not.

\begin{figure}[!htb]
    \subfloat[Fixed sample size effects on sets of 
              informative univariate, informative multivariate 
              and non-informative features. The sample 
              size is fixed of $1000$ instances. 
              \label{subfig:g}]{
      \includegraphics[width=0.5\textwidth]{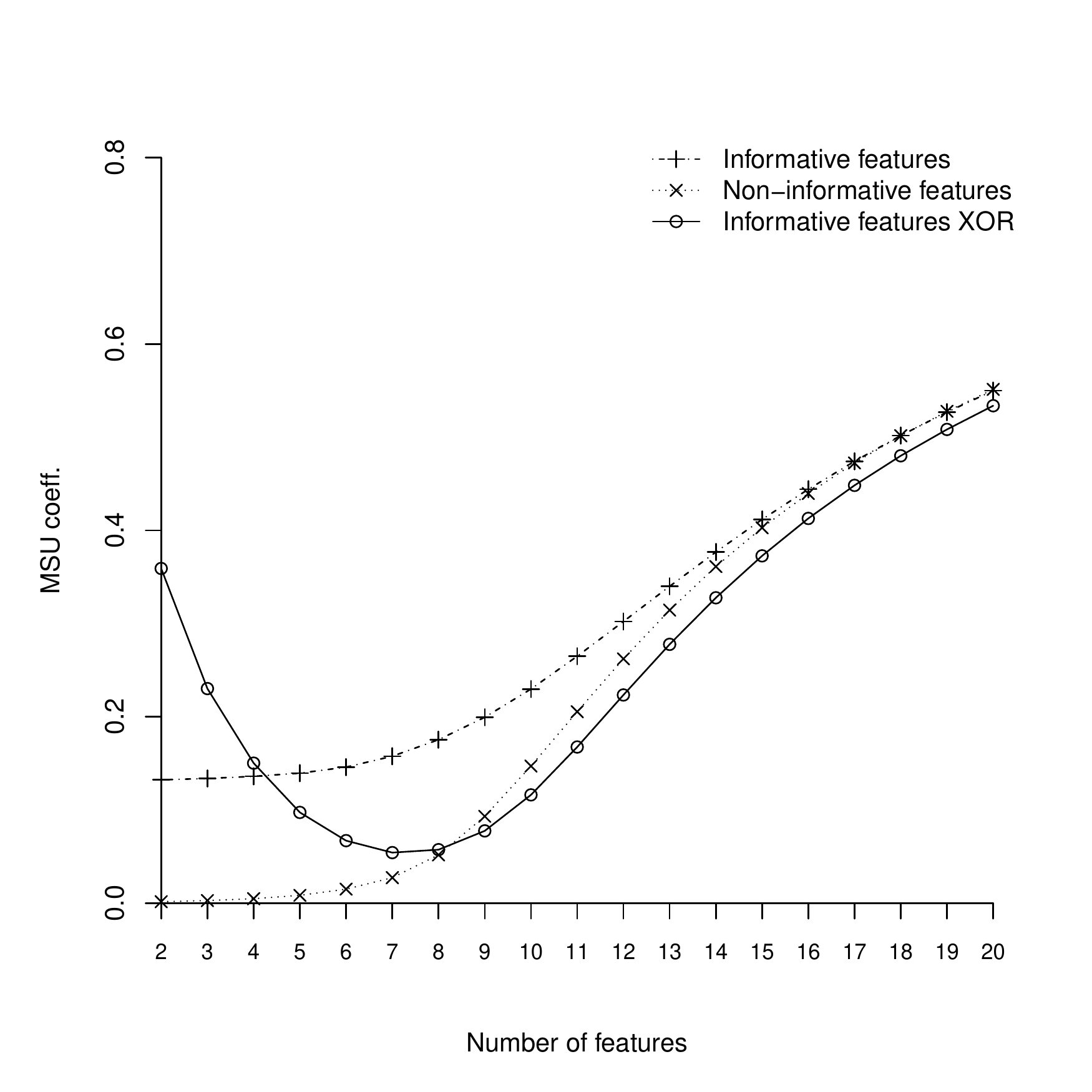}
    }
    \quad
    \subfloat[Calculated sample size effects on sets of 
              informative univariate, informative multivariate 
              and non-informative features.
              \label{subfig:h}]{
      \includegraphics[width=0.5\textwidth]{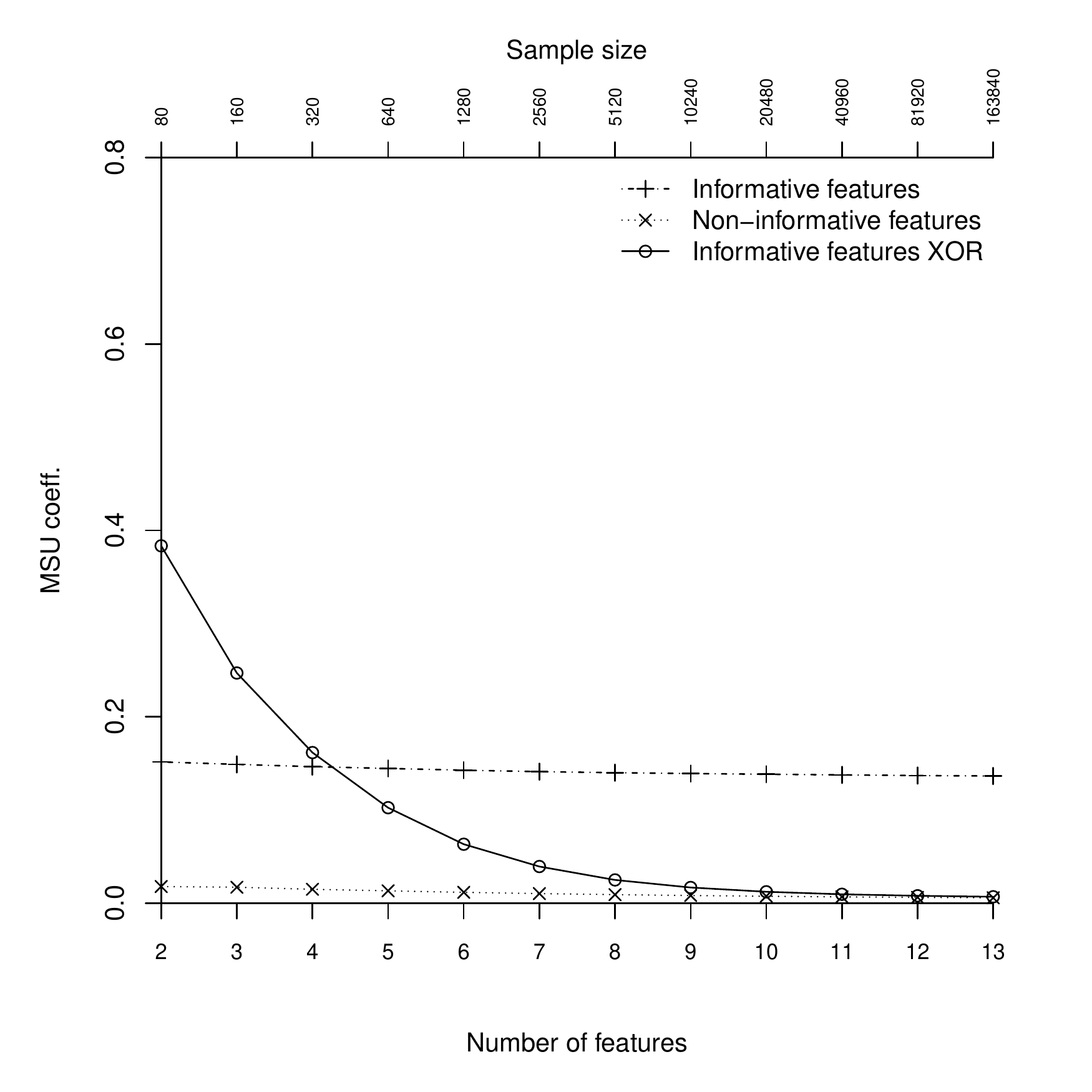}
    }
    \caption{The effects of varying the number of features with
             fixed and calculated sample size on MSU.
             The univariate cardinality is $2$.}
    \label{fig:figuras-gh}
\end{figure}

\textit{Controlling the behavior of MSU}
 Let us consider 
the above results on sample size again. The effects of 
varying sample size are shown in Figures 
\ref{fig:figuras-ef}(a) and \ref{fig:figuras-ab}(b).
We can see that a 
reasonable MSU behavior occurs when sample 
size is approximately equal to a function of the 
multivariate cardinality, given by:
{\small
\begin{equation}
Sample\ size \approx 10\left\vert{class}\right
\vert\prod_{i=1}^{n} \left\vert{f_{i}}\right\vert.
\end{equation}}
The results of arbitrary versus calculated sample sizes 
as proposed are illustrated in Figures 
\ref{fig:figuras-ef} and \ref{fig:figuras-gh}. 
The irrelevant features cause no significant bias 
in the latter, which is the desired pattern. 
For the set of informative features with high univariate 
cardinality, Figure \ref{fig:figuras-ef} (b) 
shows that the MSU decreases in an exponential-like 
shape as in the univariate case. 
However, with low univariate cardinality, 
Figure \ref{fig:figuras-gh} (b) shows the desired pattern. 
For the experiment with interactions, 
Figure \ref{fig:figuras-gh}(b) illustrates an expected 
pattern, since as we increase the number of features 
the likelihood of having a combination of the $XOR$ 
type approaches zero.

\quad

\textbf{5. Conclusions}
\smallskip


In this paper, we have considered the bias problem present 
in the MSU measure in the context of feature selection. 
We have established that the factors associated to bias 
in the detection of interactions and group correlations 
among different features are the univariate cardinality, 
the multivariate cardinality and the sample size.

Given a data set, the values of these factors are 
known {\it a priori}. We propose an empirical relationship 
between the factors, allowing the development of criteria 
for the conformation of feature subsets to be evaluated 
via the MSU as part of a feature selection process. 
In all tested cases the relationship allows to determine 
the condition where the measure has a controlled bias.

At the moment, we are studying the performance of MSU 
under known data densities found in practice. 
Furthermore, the MSU behavior should be analyzed on 
high dimensional real datasets from several domains. 

\quad

{\bf Acknowledgments}. C.E.S. acknowledges PRONII-CONACyT-Paraguay.

\end{document}